\pdfoutput=1

\documentclass[11pt]{article}

\usepackage[final]{acl}

\usepackage{times}
\usepackage{latexsym}

\usepackage[T1]{fontenc}

\usepackage[utf8]{inputenc}

\usepackage{microtype}

\usepackage{inconsolata}

\usepackage{graphicx}

\usepackage{booktabs}
\usepackage{amsmath}
\usepackage{makecell}
\usepackage{subcaption}
\usepackage[english]{babel}
\usepackage{xcolor}
\usepackage{color,colortbl}

\usepackage{microtype}
\usepackage{hyperref}
\usepackage{url}
\usepackage{booktabs}
\usepackage{amsmath,amssymb}
\usepackage{makecell}
\usepackage{float}
\usepackage{wrapfig}
\usepackage{pifont}
\usepackage{subcaption}
\usepackage{enumitem}
\usepackage{tabularx}
\usepackage{comment}

\title{A Practical Analysis of Human Alignment with *PO}

\author{
  Kian Ahrabian\textsuperscript{1}\thanks{Work done during an internship at Microsoft.} \\
  \And
  Xihui Lin\textsuperscript{2} \\
  \And
  Barun Patra\textsuperscript{2} \\
  \AND
  Vishrav Chaudhary\textsuperscript{2} \\
  \And
  Alon Benhaim\textsuperscript{2} \\
  \And
  Jay Pujara\textsuperscript{1} \\
  \AND
  Xia Song\textsuperscript{2} \\
  \AND
  \vspace{-0.65cm} \\
  \textsuperscript{1}University of Southern California, Information Sciences Institute \\
  \textsuperscript{2}Microsoft \\
  \texttt{ahrabian@usc.edu,\{xihlin,barun.patra@microsoft.com\}} \\
  \texttt{\{vchaudhary,alonbenhaim\}@microsoft.com,jpujara@isi.edu,xiaso@microsoft.com} \\
}

\begin{document}
\maketitle

\begin{abstract}
At the forefront of state-of-the-art human alignment methods are preference optimization methods (*PO).
Prior research has often concentrated on identifying the best-performing method, typically involving a grid search over hyperparameters, which can be impractical for general practitioners.
In this paper, we examine the robustness of existing state-of-the-art methods to varying hyperparameters in a realistic out-of-distribution (OOD) scenario that mirrors real-world applications of human alignment.
Our goal is to empirically find the method that increases the likelihood of achieving better results through the lens of various metrics, such as KL divergence and response length.
We also introduce LN-DPO, a simple length-normalized version of DPO that is more stable across hyperparameters, effectively reduces the average response length, and improves performance.
Our analysis of state-of-the-art reference-free (\textit{i.e.,} SimPO) and reference-dependent (\textit{i.e.,} DPO and LN-DPO) methods reveals that they perform similarly at their peak (\textit{i.e.,} best possible scenario).
However, we uncover that the pattern of change in performance greatly varies as we move away from the best possible scenario.
\end{abstract}

\section{Introduction}

In recent years, the quality of large language models (LLMs) has been constantly increasing~\cite{chiang2024chatbot}, achieving impressive results across tasks and benchmarks~\cite{abdin2024phi,llama3modelcard,achiam2023gpt,team2023internlm,qwen2}.
However, even with the most rigorous filtering heuristics, the training data~\cite{together2023redpajama,penedo2024finewebdatasetsdecantingweb} is typically contaminated with undesirable content that can lead to unacceptable behaviors~\cite{bender2021dangers,gehman2020realtoxicityprompts}.
To improve the model's alignment with human preferences, the de-facto approach has been to learn from human/AI-generated preference data (\textit{e.g.,} a chosen and a rejected response for each prompt).
In particular, off-policy preference optimization methods (*PO) have been prevalent given their good performance and ease of implementation ~\cite{rafailov2024direct,hong2024reference,meng2024simpo}. 

\begin{table}[t]
    \centering
    \resizebox{\columnwidth}{!}{
    \begin{tabular}{l|c|ccc}
        \toprule
        & \textbf{DPO} & \textbf{LN-DPO} & \textbf{SimPO} \\
        \midrule
        \textbf{Mean Score} & 1.6 & +0.3\% & \underline{+2.7}\% \\
        \midrule
        \textbf{Mean Length} & 119.8 & -15.9\% & \underline{-22.9}\% \\
        \midrule
        \textbf{KL Divergence} & 55.0 & \underline{-26.0}\% & -20.7\% \\
        \midrule
        \textbf{Win vs. Chosen} & 77.1\% & +0.8\% & \underline{+3.1}\% \\
        \midrule
        \textbf{Win vs. SFT} & 60.7\% & +2.1\% & \underline{+5.0}\% \\
        \bottomrule
    \end{tabular}
    }
    \caption{\textbf{Best *PO Performance}. The metrics are normalized by the respective DPO performance. The underlined values indicate the best performance.}
    \label{fig:best}
    \vspace{-0.4cm}
\end{table}

\begin{table*}[t]
    \centering
    \begin{tabular}{lll}
        \toprule
        \textbf{Method} & \textbf{Objective} & \textbf{Hyperparameters} \\
        \midrule
        DPO & $-\log \sigma \left(\beta \log \frac{\pi_\theta (y_w | x)}{\pi_\text{ref} (y_w | x)} - \beta \log \frac{\pi_\theta (y_l | x)}{\pi_\text{ref} (y_l | x)}\right)$ & $\beta \in \{ 0.01, 0.05, 0.1, 0.3, 0.5 \}$ \\
        \midrule
        SimPO & $-\log \sigma \left(\frac{\beta}{|y_w|} \log \pi_\theta (y_w | x) - \frac{\beta}{|y_l|} \log \pi_\theta (y_l | x) - \gamma\right)$ & \makecell[l]{$\beta \in \{ 1.0, 1.5, 2.0, 2.5 \}$\\$\gamma \in \{ 0.5, 0.8, 1.0, 1.2, 1.4, 1.6 \}$} \\
        \midrule
        LN-DPO & $-\log \sigma \left(\frac{\beta}{|y_w|} \log \frac{\pi_\theta (y_w | x)}{\pi_\text{ref} (y_w | x)} - \frac{\beta}{|y_l|} \log \frac{\pi_\theta (y_l | x)}{\pi_\text{ref} (y_l | x)}\right)$ & $\beta \in \{ 1.0, 1.5, 2.0, 2.5, 3.0, 3.5 \}$ \\
        \bottomrule
    \end{tabular}
    \caption{\textbf{*PO Optimization Objectives.}
    The preference data is formulated as $D = (x, y_w, y_l)$, where $x$ is the prompt and $y_w$ and $y_l$ are the chosen and rejected responses.
    }
    \label{tab:obj}
    \vspace{-0.2cm}
\end{table*}

One commonly occurring practice when reporting the performance of new methods is to compare their best-performing variant (after a hyperparameter grid search) to a default baseline with a fixed set of hyperparameters.
However, from a practical perspective for future users, these comparisons do not provide a good answer to the problem of which method is expected to achieve higher performance, given a fixed budget for hyperparameter search, as doing broad grid searches is often computationally infeasible for many practitioners.
To this end, in this work, we aim to empirically identify the more robust method to hyperparameter variations while still being competitive in performance.

We set up our experiments in a realistic out-of-distribution (OOD) setting, focused on safety and helpfulness domains, where the train and test datasets share a common core goal, but their samples are generated from different distributions (\textit{e.g.,} AI and human expert).
This setting resembles real-world scenarios as it simulates the release of large generative models for public use.
Moreover, to better understand the behavior of the state-of-the-art models, we take the best-performing reference-free and reference-dependent models (as reported by~\citet{meng2024simpo}) and analyze them through the lens of standard metrics such KL divergence, response length, and win rate.
We also introduce an embarrassingly simple length-normalized extension of vanilla Direct Preference Optimization (DPO)~\cite{rafailov2024direct}, LN-DPO, that effectively mitigates the issue of lengthy generations without any apparent performance degradation\footnote{Concurrently, \citet{meng2024simpo} have added a similar method to their experiments (updated on July 7th, 2024). Here, we present a more thorough analysis and comparison.}.
In summary, our contributions are as follows:
\begin{itemize}[itemsep=0pt, topsep=2pt]
    \item We examine state-of-the-art reference-free and reference-dependent preference optimization methods across a wide range of hyperparameters in a real-world setup.
    \item We analyze the performance of these methods on critical metrics such as mean response length, mean score on a gold reward model, win rate vs. chosen and SFT, and KL vs. SFT.
    \item We introduce and examine LN-DPO, a simple length-normalized version of DPO that is more stable across hyperparameters, effectively reduces the average response length and improves performance.
\end{itemize}

\section{Related Work}

\begin{figure*}[t]
\centering
  \includegraphics[width=0.95\textwidth]{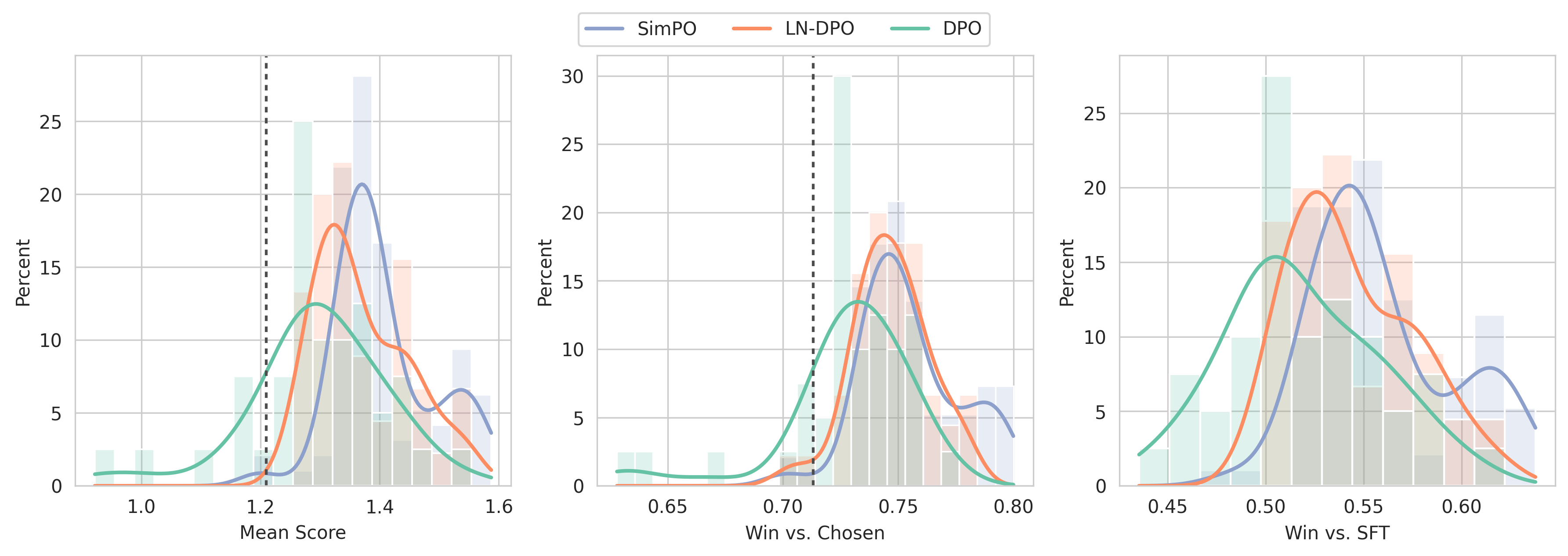}
    \caption{
        \textbf{*PO Performance Distribution}.
        Each sample in the distribution represents the performance of one set of hyperparameters on the denoted metric. The dashed line indicates the performance of the initial SFT model.
    }
    \label{fig:perf}
    \vspace{-0.2cm}
\end{figure*}

Since the introduction of DPO~\cite{rafailov2024direct}, there has been a body of works with new optimization objectives improving the performance and efficiency~\cite{azar2024general,tang2024generalized,hong2024reference,rosset2024direct,meng2024simpo,xu2024contrastive,ethayarajh2024kto}.
These methods can be partitioned into two groups: reference-free~\cite{meng2024simpo,hong2024reference} and reference-dependent~\cite{rafailov2024direct,park2024disentangling}.
Reference-free methods generally benefit from fast training runs, while reference-dependent methods have terms baked into their objective to control divergence from the reference model. 
In this work, we compare SimPO~\cite{meng2024simpo}, a recent state-of-the-art reference-free method, with DPO and LN-DPO as reference-dependent methods (see \autoref{app:related} for extended related work).

\section{Experimental Setup}

\subsection{Datasets}
For our datasets, we follow the setup introduced by~\citet{xu2024dpo}.
Specifically, we use the double safe/unsafe filtered train subset of SafeRLHF~\cite{safe-rlhf} for training and the test subset of HH-RLHF~\cite{ganguli2022red} for evaluation.
This setup closely resembles real-world scenarios where even though models are trained on various domains (\textit{e.g.,} safety and helpfulness in our experiments), they have to generalize to similar unseen queries while interacting with the users.

\begin{figure*}[t]
  \includegraphics[width=\textwidth]{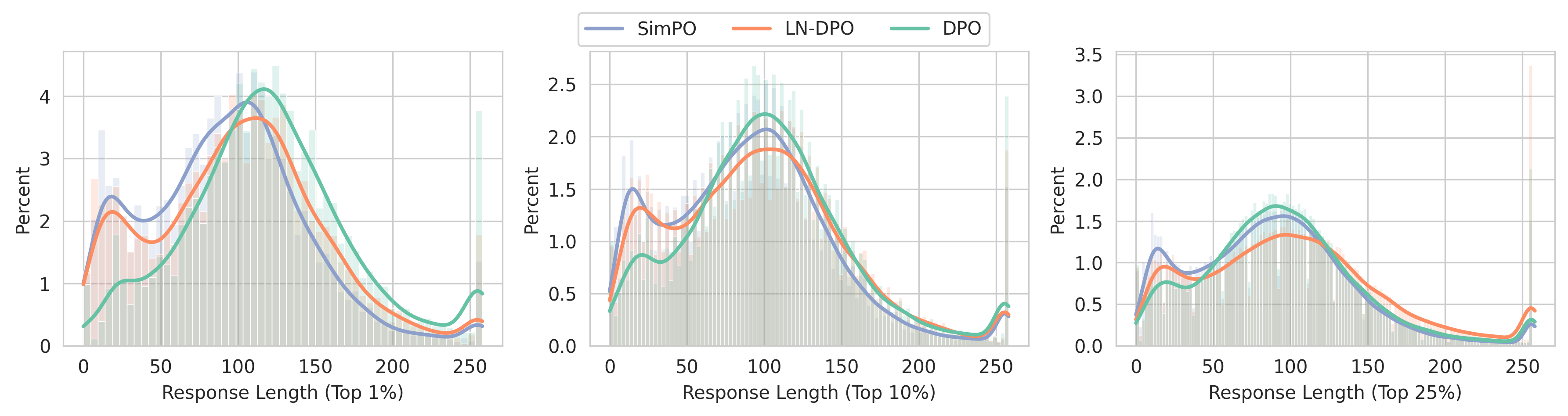}
  \caption{\textbf{Response Length}.
  The top k\% ($k \in \{1,10,25\}$) denotes the percentage of best-performing hyperparameters taken from each method's runs.}
  \label{fig:len}
  \vspace{-0.2cm}
\end{figure*}

\begin{figure*}[t]
  \includegraphics[width=\textwidth]{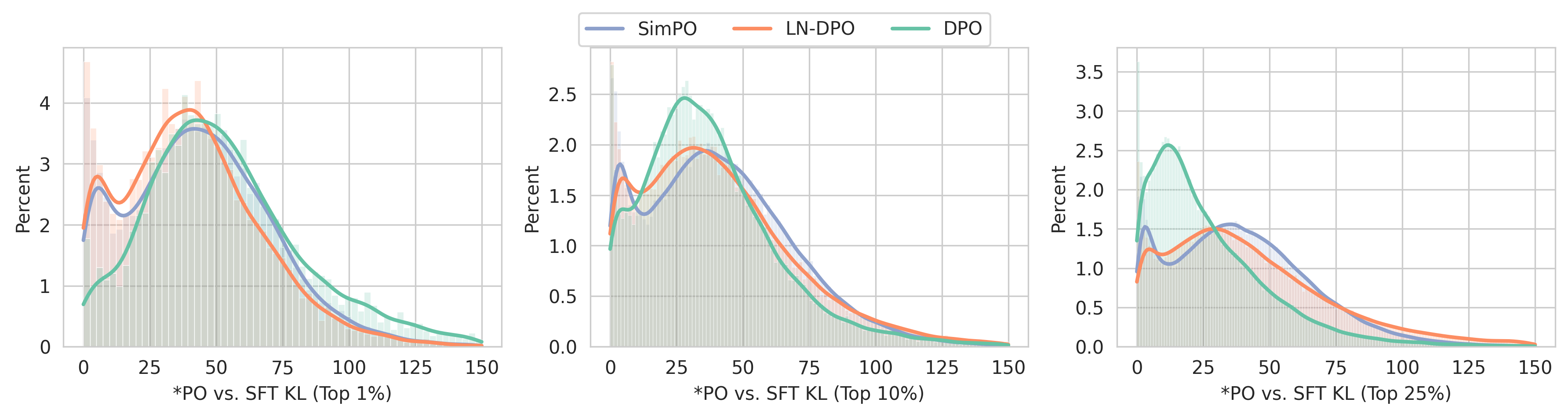}
  \caption{\textbf{KL Divergence}.
  The top k\% ($k \in \{1,10,25\}$) denotes the percentage of best-performing hyperparameters taken from each method's runs.}
  \label{fig:kl}
\end{figure*}

\subsection{Models}
For all our experiments, we chose the Phi-3 Medium model~\cite{abdin2024phi} due to its high performance across benchmarks and small size, ensuring computational tractability.
To evaluate the trained models, we use the OpenAssistant reward model~\cite{kopf2024openassistant} to score the quality of their generated responses.
We chose this model due to its small size and use in prior works~\cite{xu2024dpo}, ensuring fast and correct evaluations.

\subsection{Optimization Objectives}
Considering the performances reported by~\citet{meng2024simpo}, we choose DPO as our reference-dependent method and SimPO as our reference-free method.
While DPO has an implicit length normalization through the reference model, the variance of the reward (\textit{i.e.,} $\log \frac{\pi_{\theta}}{\pi_{\text{ref}}}$) increases with response length.
As such, inspired by explicit length regularization in SimPO and R-DPO~\cite{park2024disentangling}, we further normalize it with the response length similar to SimPO, which we call LN-DPO (see \autoref{app:ln_dpo_simpo_connection} for more details).

\subsection{Connection between LN-DPO and SimPO}
\label{app:ln_dpo_simpo_connection}
LN-DPO is similar to an adaptive margin version of SimPO with per sample margin defined as
\begin{equation}
    \gamma_{w,l} = \log\left.\frac{\pi_{\text{ref}}(y_w|x)}{|y_w|}\right. - \log\left.\frac{\pi_{\text{ref}}(y_l|x)}{|y_l|} \right. .   
\end{equation}
Essentially, this adaptive margin encourages larger margins for pairs with large margins in the reference policy.
Depending on the quality of the reference model and the labels, this change could be beneficial compared to SimPO's constant margin.
The adaptive margin focuses more on "easier" pairs (\textit{i.e.,} pairs that have some prior evidence to be different) while less on "harder" pairs (\textit{i.e.,} pairs that are closer), which means that LN-DPO is potentially less prone to overfitting and less sensitive to wrong labels.

\section{Training Regimen}

\label{setup:train}
Following the common practice, before the preference optimization step we do a supervised fine-tuning (SFT) step.
Specifically, we first run a grid search over the following hyperparameters: epochs $\in \{1, 3\}$ and learning rate $\in \{1e-6, 3e-6, 1e-5, 2e-5\}$.
Then we evaluate the final checkpoints against the test set and choose the one with the highest performance.
This procedure ensures that the preference optimization methods are initialized from a good checkpoint.
For the preference optimization methods, we run a grid search using 1) the same ranges as SFT for epochs and learning rate and 2) common values for method-specific hyperparameters as used in prior works~\cite{meng2024simpo,rafailov2024direct,hong2024reference}.
\autoref{tab:obj} presents the method-specific ranges used in our experiments.
In all of our experiments, the batch size is set to 256.

\section{Metrics}
Our analysis focuses on the following five metrics:
\begin{itemize}[itemsep=0pt,topsep=0pt]
    \item {\bf Mean Score: } The average score of the generated responses, as judged by the gold reward model.
    \item {\bf Win vs. Chosen: } The fraction of samples where the gold reward model assigns a higher score to the generated response compared to the chosen response in the dataset.
    \item {\bf Win vs. SFT: } The fraction of samples where the gold reward model scores the generated response higher than the initial SFT model's response.
    \item {\bf KL divergence: } The summed difference of log probabilities between the SFT and the trained models over the samples.
    \item {\bf Response length: } The number of tokens in the generated response under the tokenization space of the base model.
\end{itemize}

\section{Implementation Details}
We generate all the responses by sampling with a \texttt{temperature $= 0.7$}, and \texttt{top\_p $= 0.95$}.
Moreover, \texttt{max\_generation\_length} is set to 256 across all experiments, following the setup by~\citet{xu2024dpo}.
All our experiments are carried out on a cluster with 256$\times$A100 80GB GPUs.
Finally, we implemented our code using the Transformers~\cite{wolf-etal-2020-transformers}, TRL~\cite{vonwerra2022trl}, and PyTorch~\cite{paszke2019pytorch} libraries.

\section{Experimental Results}

\subsection{Hyperparameter Robustness}

\paragraph{Best Performance.}
Following the common practice, we compare the best performance achieved by each method in \autoref{fig:best}.
As evident, at their peaks, SimPO, LN-DPO, and DPO score similarly (within a 0.05 point on average).
However, SimPO and LN-DPO show an edge in terms of the rest of the metrics.
Specifically, we can observe the effectiveness of the length normalization term.
We also notice a significant decrease in KL divergence.
However, KL for SimPO decreases less than LN-DPO, showcasing a more significant divergence from SFT.
For more details on tuning these models, see \autoref{app:hyp_sens}.

\begin{table}[t]
    \centering
    \small
    \begin{minipage}{\columnwidth}
        \centering
        \begin{tabular}{l|cccc}
            \toprule
            \makecell[c]{\textbf{\%}} & \textbf{DPO} & \textbf{LN-DPO} & \textbf{SimPO} \\
            \midrule
            \textbf{DPO} & - & 49.04 & 47.51 \\
            \midrule
            \textbf{LN-DPO} & \underline{49.47} & - & 46.43 \\
            \midrule
            \textbf{SimPO} & \underline{51.12} & \underline{51.09} & - \\
            \bottomrule
        \end{tabular}
        \subcaption{Best}
        \label{tab:h2h:top1}
        \vspace{0.2cm}
    \end{minipage}
    \begin{minipage}{\columnwidth}
        \centering
        \begin{tabular}{l|cccc}
            \toprule
            \makecell[c]{\textbf{\%}} & \textbf{DPO} & \textbf{LN-DPO} & \textbf{SimPO} \\
            \midrule
            \textbf{DPO} & - & 45.72 & 44.33 \\
            \midrule
            \textbf{LN-DPO} & \underline{51.77} & - & 47.28 \\
            \midrule
            \textbf{SimPO} & \underline{54.34} & \underline{50.13} & - \\
            \bottomrule
        \end{tabular}
        \subcaption{75th Percentile}
        \label{tab:h2h:top2}
    \end{minipage}
    \caption{\textbf{Head-to-head *PO Comparison.}
    Each cell represents the win rate of the row method over the column method.
    The underlined values indicate the row method beating the column method.}
    \label{tab:h2h}
    \vspace{-0.2cm}
\end{table}

\paragraph{Head-to-head Performance.}
While comparing the pure performances achieved on the desired metrics is usually good enough to contrast different methods, there are potential cases where the averaging could be exploited (\textit{e.g.,} outliers with high rewards).
Hence, it is essential also to do a head-to-head per sample comparison, which provides more fine-grained insights.
\autoref{tab:h2h} compares each method's best and 75th percentile performance.
Notably, we observe a sharp performance drop in DPO from the best to the top 25\% model, in contrast to the other two.
This occurrence highlights the practical flaw in only comparing the best performances.

\paragraph{Expected Performance.}
Given the limited resources that most users have, it is extremely difficult to run broad hyperparameter searches to find the best-performing combination.
As such, it becomes crucial to analyze hyperparameter robustness, which provides insights into the expectation of finding good hyperparameters set from a limited search.
\autoref{fig:perf} presents the performance distribution *PO methods following a grid search over the hyperparameters denoted in \autoref{tab:obj} and \autoref{setup:train}.
As evident, SimPO and LN-DPO effectively increase the average performance (i.e., shifting the distributions to the right) across hyperparameters, showcasing their superiority.
Note that we stretched the range of hyperparameters until a plateau or an extreme variance was observed.

\subsection{Response Length}
Since length exploitation is a critical issue ~\cite{park2024disentangling}, we compare the response lengths across samples generated by the top k\% ($k \in \{1,10,25\}$) of each method's best-performing hyperparameters.
As illustrated in \autoref{fig:len}, on the best set of hyperparameters (\textit{i.e.,} top 1\%), the non-DPO methods showcase a left shift in length distribution (compared to DPO), which is a desired effect.
However, this phenomenon starts to diminish as we include worse-performing hyperparameters.
For example, LN-DPO has a higher rate than DPO in the tail-end of the top 25\% distribution.
Overall, we observed that both length-normalized models perform superior to DPO, with SimPO producing the shortest responses across the distribution.

\subsection{KL Divergence (vs. SFT)}
Since reference-free methods are not normalized against a reference policy (\textit{e.g.,} the SFT model), reward hacking might occur (\textit{i.e.,} lower loss with degraded performance).
Therefore, we compare the KL divergence in \autoref{fig:kl} across samples generated by the top k\% ($k \in \{1,10,25\}$) of each method's best-performing hyperparameters.
As evident, both SimPO and LN-DPO achieve lower KLs at their peak.
However, as we move toward worse-performing models, DPO achieves lower KL (at 10\%).
This phenomenon is due to many DPO runs failing to learn beyond the SFT model.



\section{When to use LN-DPO over SimPO?}
\label{app:ln_dpo_vs_simpo}
While SimPO achieves superior performance on most metrics compared to LN-DPO, the lack of a reference policy regularization could lead to drastic divergence from the initial checkpoint, as also shown in our experiments.
This issue then could cause a degradation of performance on other benchmarks, which is a critical pitfall (as also observed in \citet{korbak-etal-2022-rl}).
As such, we believe there are various scenarios where LN-DPO should be preferred to SimPO.
We leave further experiments over this direction to future works.

\section{Conclusion}
In this work, we introduce LN-DPO, a length-normalized variation of DPO that reduces the average response length while staying reference-dependent.
Moreover, we present a thorough analysis of LN-DPO and two state-of-the-art reference-dependent and reference-free preference optimization methods in a simulated real-world scenario for safety and helpfulness domains.
Specifically, we cover the behavior of these methods across a wide range of hyperparameters under metrics such as mean response length, KL divergence (vs. SFT), and win rate (vs. chosen and SFT).
Our experiments showcase state-of-the-art methods' strengths and weaknesses and provide insights for other practitioners.

\section*{Limitations}
Due to the extremely high costs of running such experiments (\textit{i.e.,} roughly 86000 GPU hours for the current experiments), in this work, we only experimented with a small set of models, methods, and datasets.
While this might limit generalizability, we believe the existence of such analysis is critical to help practitioners save costs.
Moreover, since the conclusion of our experiments, new reward models with higher performance have been released (\textit{e.g.,} ArmoRM~\cite{wang2024interpretable}); however, we still rely on older, smaller models to keep the evaluation tractable on such a high number of runs.

\section*{Acknowledgements}
This work was partially funded by the Defense Advanced Research Projects Agency with the award HR00112220046.

\bibliography{main}

\appendix

\section{Extended Related Work}
\label{app:related}

\paragraph{Online Algorithms.}
Reinforcement learning from human/AI feedback (RLHF/RLAIF) is among the common approaches for aligning LLMs to human preferences~\cite{christiano2017deep,bai2022training,stiennon2020learning,bai2022constitutional}, and has been used to train models such as GPT-4~\cite{achiam2023gpt} and Llama-3~\cite{llama3modelcard}.
In most cases, these approaches are comprised of three stages: 1) supervised fine-tuning~\cite{taori2023stanford,zhou2024lima,xia2024less}, 2) reward modeling~\cite{gao2023scaling,chen2024odin,lightman2023let}, and 3) policy optimization~\cite{schulman2017proximal}.
The prominent method for policy optimization is Proximal Policy Optimization (PPO), an online on-policy approach~\cite{schulman2017proximal}.
While PPO has shown promising performances~\cite{stiennon2020learning,ouyang2022training,achiam2023gpt}, it suffers from problems such as having too many subtle details for reproducibility~\cite{huang2024thenimplementation}, 2) taking a long time for training~\cite{huang2024n+}, and 3) reward over-optimization~\cite{skalse2022defining}.

\paragraph{Offline Algorithms.}
To address the drawbacks of RLHF/RLAIF, recent works have proposed simpler and more efficient offline algorithms, particularly Direct Preference Optimization (DPO)~\cite{rafailov2024direct}, which is based on the Bradley-Terry model~\cite{bradley1952rank}.
These offline algorithms directly optimize an objective on the preference data with an implicit reward model without needing to have separate stages.
Some recent works have focused on making a broad comparison between PPO and DPO. Specifically, they showcase the potential for PPO with a gold reward model ($\sim +10\%$) while underlying the similarity to DPO  ($\sim +1\%$ averaged across benchmarks) when trained on the same data~\cite{ivison2024unpacking,xu2024dpo}.



\begin{figure*}[t]
  \includegraphics[width=\textwidth]{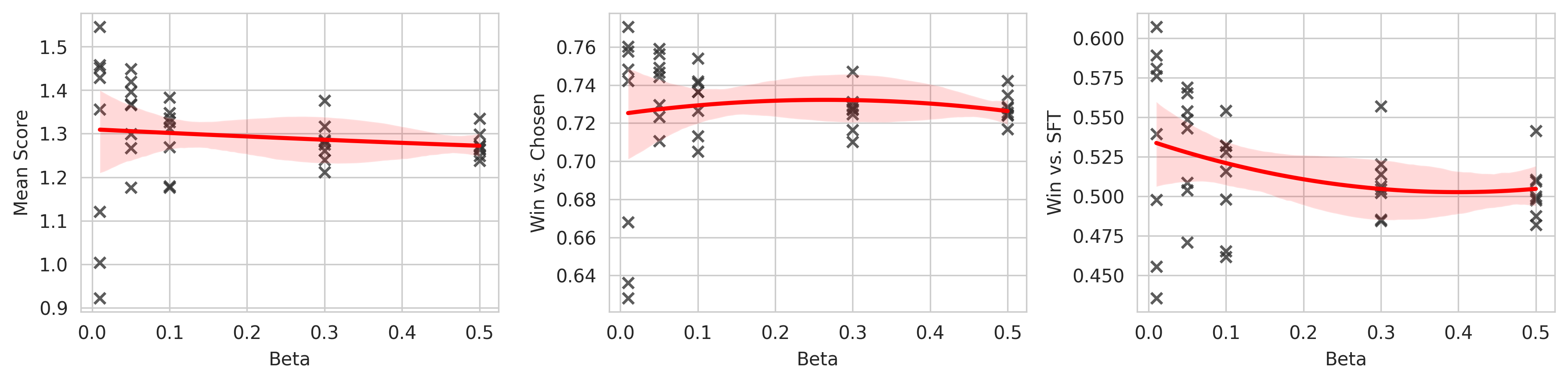}
  \caption{\textbf{DPO $\beta$}. Each point indicates a run with the corresponding $\beta$ value.}
  \label{fig:dpo}
\end{figure*}

\begin{figure*}[t]
  \includegraphics[width=\textwidth]{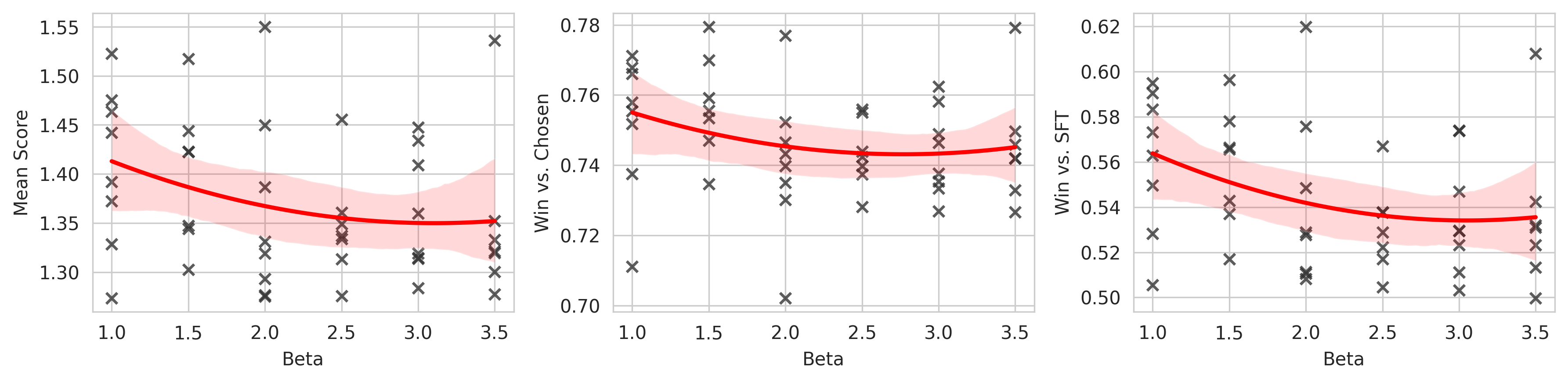}
  \caption{\textbf{LN-DPO $\beta$}. Each point indicates a run with the corresponding $\beta$ value.}
  \label{fig:dpo_ln}
\end{figure*}

\section{Hyperparameter Tuning Considerations}
\label{app:hyp_sens}
\paragraph{DPO.}
As presented in \autoref{fig:dpo}, lower $\beta$ leads to higher performances; however, as $\beta$ decreases, the performance variance increases, which showcases the method's instability.
Overall, $\beta = 0.05$ provides the best balance of stability and performance.

\paragraph{LN-DPO.}
While we initially borrowed $\beta$'s range from SimPO~\cite{meng2024simpo}, more experiments showed benefits in further decreasing its value.
\autoref{fig:dpo_ln} presents the performance spread across different runs.
From these experiments, $\beta \in [1.0, 2.0]$ contains most of the best-performing models.
Moreover, we observe the relatively low (compared to DPO) variance across the performances, showcasing another benefit of LN-DPO.

\begin{figure*}[t]
  \includegraphics[width=\textwidth]{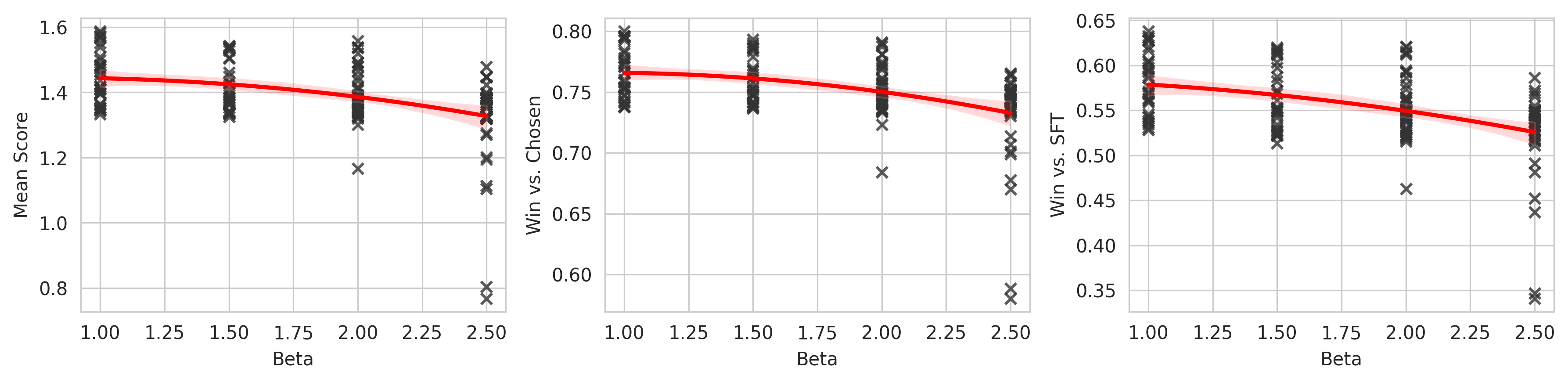}
  \caption{\textbf{SimPO $\beta$}. Each point indicates a run with the corresponding $\beta$ value.}
  \label{fig:simpo:beta}
\end{figure*}

\begin{figure*}[t]
  \includegraphics[width=\textwidth]{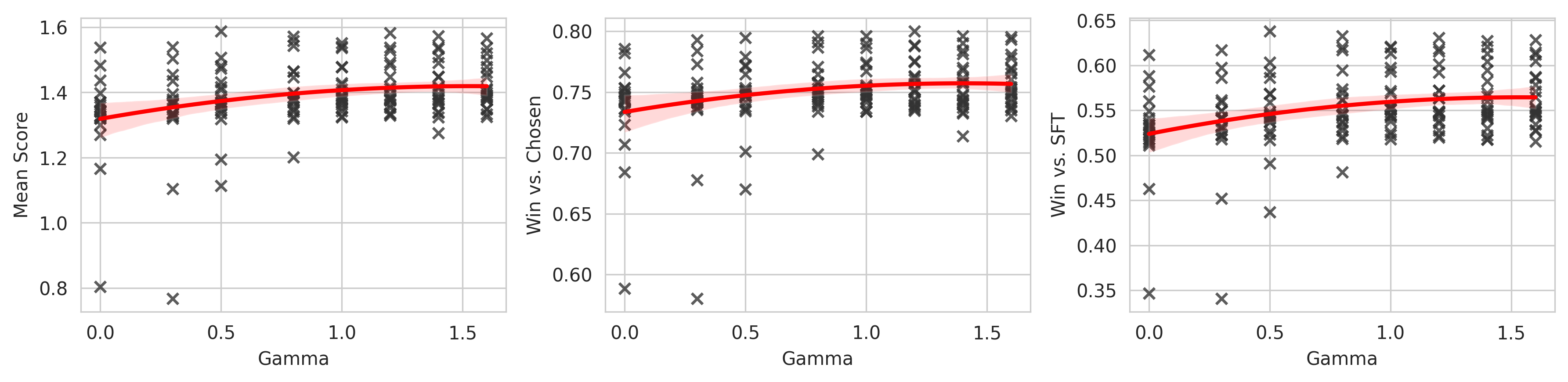}
  \caption{\textbf{SimPO $\gamma$}.  Each point indicates a run with the corresponding $\gamma$ value.}
  \label{fig:simpo:gamma}
\end{figure*}

\paragraph{SimPO.}
In contrast to the other two methods, SimPO has two method-specific hyperparameters: $\beta$ and $\gamma$.
As illustrated in \autoref{fig:simpo:beta}, on average, lower $\beta$ values lead to better performance.
We believe the performance uptick in the lower range is due to a difference in the average length of this work's and the original work's training sets.
Moreover, as showcased in \autoref{fig:simpo:gamma}, the best performing models have a $\gamma \in [1.0, 1.4]$, in line with the suggestion by~\citet{meng2024simpo}.
Notably, $\beta$ and $\gamma$ have a relatively low variance across experiments, another upside of SimPO.

\section{The Answer to the Ultimate Question}
Based on our collective empirical results, we believe SimPO to be the best starting point among the three methods, mainly due to its robustness toward hyperparameter variations and effective length reduction.
As for SimPO's hyperparameters, we recommend $\beta \in \{1.0, 1.5\}$ and $\gamma \approx 1.2$.
Moreover, while LN-DPO is consistently second-best in most of our experiments, we discuss scenarios for choosing it over SimPO in \autoref{app:ln_dpo_vs_simpo}.

\end{document}